\documentclass[10pt,twocolumn]{article}

\usepackage[margin=0.75in]{geometry}
\usepackage{graphicx}
\usepackage{float}
\usepackage{booktabs}
\usepackage{amsmath,amssymb}
\usepackage{xcolor}
\usepackage{tikz}
\usetikzlibrary{arrows.meta,positioning,calc}
\usepackage[hidelinks]{hyperref}
\usepackage{microtype}
\title{\vspace{-1.5em}
Massive Activations Are Architecturally Robust:\\
A Controlled Scratch/Commitment Residual Stream Test\vspace{-0.5em}}

\author{
Maruthi Vemula\\
University of North Carolina at Chapel Hill\\
\texttt{vemula@unc.edu}
}
\date{}

\begin{document}
\maketitle

\begin{abstract}
\noindent
Trained transformers reliably develop massive activations, a small number of hidden
dimensions whose magnitude is far above the median and which concentrate on the
sequence-start token. Whether these outliers are a removable artifact of the residual
stream's overloaded read and write role, or instead a functional necessity, is actively
debated. We test the artifact hypothesis directly, with an architectural intervention. Our
architecture, Ledger Residuals, splits the residual stream into a mutable scratch stream
(Deliberation) that intermediate computation may freely overwrite and a protected,
decode-only accumulator (Commitment) that holds the representation the model reads out. If
massive activations exist only because one stream is forced to be both scratchpad and
answer, then a dedicated answer channel should remove the need for them. We find that it
does not. In matched-loss language models at the 160M and 290M scales, the model rebuilds
the canonical fixed-dimension, start-token outlier inside the protected channel. The
rebuilt feature is smaller in magnitude than in a standard transformer but more sharply
concentrated on the start token, and a stronger sparsity penalty makes it more persistent
and more concentrated still, rather than removing it. Massive activations therefore look
architecturally robust: they re-emerge in whichever representation the model decodes from,
which is what we would expect if they are functional rather than incidental. We release our
architecture and measurement code.
\end{abstract}

\section{Introduction}
Trained transformers reliably develop a striking and counterintuitive feature. A small number
of hidden-state coordinates take on magnitudes thousands of times larger than the median, and
they do so in a remarkably structured way: the same few feature dimensions are involved across
inputs, and the large values land on just a few token positions, most often the sequence-start
or delimiter token [1]. These massive activations appear within the first few layers, stay
almost constant through the rest of the network, and behave like input-independent bias terms,
and they are load-bearing in the strongest sense, since zeroing out even the largest few makes
the model collapse [1]. They are accompanied by attention sinks, where a disproportionate share
of attention mass is routed to those same initial tokens regardless of what the tokens contain
[2].

Why transformers do this is genuinely unsettled, and the literature has split into two camps.
One camp treats massive activations as an artifact, a side effect of how the architecture is
built rather than something the computation actually needs. Its evidence is that their magnitude
can be reduced at little or no cost to language modeling, through explicit key and value biases
[1], QK-normalization or gating [3], or clipped and gated softmax variants [9], and that a
recent analysis ties the joint appearance of massive activations and attention sinks to the
pre-norm design rather than to any functional requirement [3]. The other camp treats them as
functional. Attention sinks have been argued to keep representations from over-mixing as depth
grows [8] and to follow directly from the softmax sum-to-one constraint [11], massive
activations have been interpreted as adaptive regulators of gradient flow during training [10],
and sinks have been shown to be provably necessary for a class of trigger-conditional
computations [14].

We set out to test the artifact hypothesis directly, by building the architecture it implicitly
recommends and seeing whether it works. Our starting point is the standard mechanistic picture
of the residual stream as a single shared memory that every layer can read from and write to,
where a component's output lingers in some subspace until a later layer overwrites it [5]. This
forces one object to play two roles that pull against each other. It is a scratchpad, which
intermediate computation should be free to overwrite, and it is at the same time the answer,
which the unembedding reads out at the end and which therefore has to survive everything written
on top of it. Suppose massive activations are the model's improvised fix for this conflict, a
stable, content-independent channel it carves out of a stream that was never meant to hold one.
Then separating the two roles, and handing the model a clean answer channel of its own, ought to
make the fix unnecessary.

This is exactly the intervention we build. Ledger Residuals, which we describe in
Section~\ref{sec:method}, replace the single residual stream with two. The first is a mutable
Deliberation stream that sublayers may freely write to and erase from, and the second is a
protected, append-only Commitment stream that is the only channel the unembedding decodes. The
model now has a dedicated place to keep the representation it will be judged on, kept apart from
the scratch space it computes in. If the artifact hypothesis is right, a model with such a
channel should no longer need massive activations in the representation it decodes from.

The result is a clear negative. The intervention does not remove massive activations. In
matched-loss language models at the 160M and 290M scales, the model rebuilds the familiar
fixed-dimension, start-token outlier inside the protected commitment channel, the one place we
had tried to keep clean. The rebuilt outlier is smaller in magnitude than in a standard
transformer, but it is more sharply concentrated on the start token, and the obvious remedy of
pressuring the channel to commit sparsely only makes the feature more persistent and more
concentrated, not weaker. Wherever the model reads its answer from, it puts the feature there,
which we take as evidence on the functional side of the debate.

Our main contribution is this test and what it tells us, made possible by an architecture built
for the purpose. We introduce Ledger Residuals, an asymmetric factorization of the residual
stream into a scratch channel and a commitment channel that reduces exactly to a standard
pre-norm transformer at a known setting of its gates, which lets us use it as a controlled
instrument rather than as a competing model (Section~\ref{sec:method}). With it, we report a
negative result that holds at both the 160M and 290M scales at matched loss: a protected,
decode-only channel does not remove massive activations, and the channel the model decodes from
rebuilds the canonical fixed-dimension, start-token outlier (Section~\ref{sec:results}). We
further show that the natural remedy backfires, since a stronger commit-sparsity penalty
intensifies the rebuilt outlier rather than removing it, raising its dominant-dimension
persistence from 0.58 to 0.96 and its start-token concentration from 3.4 to 4.7. Taken together,
these findings weigh in on the functional side of the artifact-versus-function debate, and we
release the architecture and the measurement code\footnote{\url{https://github.com/MaruthiV/ledger-residuals}}
so that the test can be reproduced and pushed to larger scale.

\section{Related Work}
\textbf{Massive activations and attention sinks.}
Trained transformers concentrate extreme behavior in a handful of coordinates and a handful of
positions, and a growing body of work has documented it. Sun et al. [1] show that a small,
input-independent set of hidden dimensions takes on magnitudes thousands of times above the median.
These dimensions appear within the first few layers, persist almost unchanged through the rest of
the network, and act as bias terms whose removal makes the model collapse. The broader observation
that a few coordinates dominate transformer representations predates this framing: outlier
dimensions were found to disrupt BERT when ablated [16], to distort the geometry of similarity
comparisons [17], to be driven by token frequency during training [18], and to be the principal
obstacle to low-bit quantization at scale [19]. That earlier line of work already separated the
sheer magnitude of an outlier from the structural role that makes it load-bearing, a distinction our
results turn on. The attention-side counterpart is the attention sink [2], where a large share of
attention mass is sent to the first
few tokens no matter what they contain, and keeping those tokens around is what makes streaming
inference stable. The two phenomena are tightly linked, because the start token is both where
attention pools and where the largest activations sit. A related case is the confidence-regulation
neuron [4], which writes into directions the unembedding ignores so as to tune output confidence. It
is one more instance of a network setting aside part of its own representation as a dedicated control
channel, which is the same instinct our architecture tries to satisfy on purpose.

\textbf{The artifact-versus-function debate.}
Whether these outliers are incidental or necessary is the open question, and the evidence cuts both
ways. The artifact side points out that their magnitude can be suppressed cheaply. Explicit learnable
key and value biases remove them in small GPT-2 models [1], clipped and gated softmax variants shrink
them enough to allow low-bit quantization [9], and a recent analysis argues that the co-occurrence of
massive activations and attention sinks is largely a by-product of the pre-norm architecture rather
than a functional requirement [3]. An important nuance is that an off-by-one (softmax-1) modification
removes the attention sink but leaves the massive activations in place [13], which already hints that
the magnitude of an outlier and its structure are two different things, a distinction our results
turn on. The functional side reads the same phenomena differently. Attention sinks have been
explained as a way of stopping representations from over-mixing as depth grows [8] and as a direct
consequence of the softmax normalization constraint [11], massive activations have been cast as
adaptive regulators of gradient flow during training [10], and sinks have been proven necessary for a
class of trigger-conditional computations [14]. What this debate has lacked is a direct architectural
test, which is what we provide, and our test comes out on the functional side.

\textbf{Multi-stream and erasable residual streams.}
Our architecture sits within a line of work that modifies the residual stream itself, and it helps to
say what we borrow and what is new. Hyper-Connections [6] widen the residual path into several
parallel streams with learned mixing weights, but the streams are summed back together before the
output and are not given distinct read or write roles. Deep Delta Learning [7] lets each layer
selectively rewrite the residual content rather than only add to it, which is the erase capability we
adopt for our scratch stream, though it keeps a single stream and no protected accumulator. The
concurrent Dual-Stream Transformer [12] also splits the residual stream into two, but along a
different axis than we do: one stream is updated by attention and the other by the feed-forward
layers, and the model reads out from their combination. What none of these designs has is a
protected, decode-only channel, and that is precisely the ingredient our test needs. The closest idea
outside the residual stream is the register token in vision transformers [15], which adds spare tokens
that soak up high-norm artifacts. Registers live on the sequence axis rather than as a split of the
residual stream, but they share our intuition of giving the model a dedicated place to put things it
needs to set aside.

\section{Ledger Residuals}
\label{sec:method}

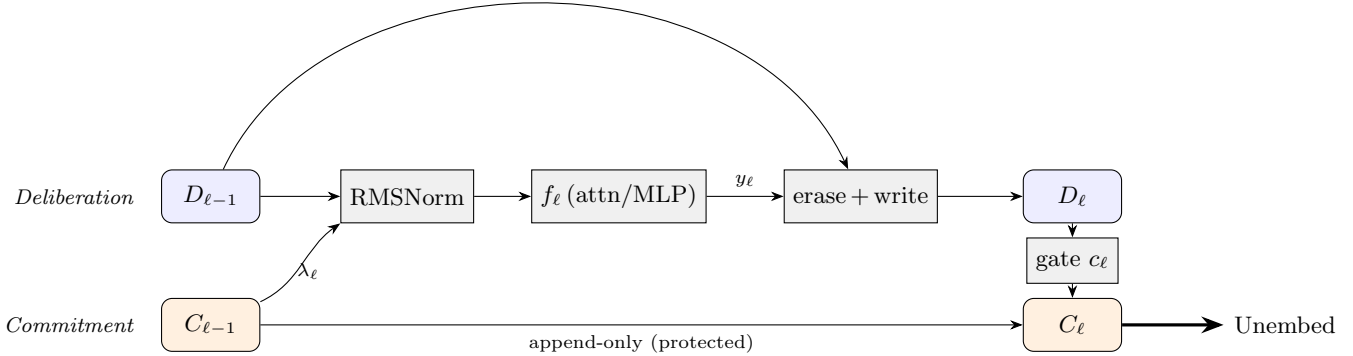
\begin{figure*}[t!]\centering
\begin{tikzpicture}[>=Stealth, font=\small,
  Dn/.style={draw, rounded corners, fill=blue!7,   minimum width=13mm, minimum height=7mm},
  Cn/.style={draw, rounded corners, fill=orange!12, minimum width=13mm, minimum height=7mm},
  op/.style={draw, fill=black!6, minimum width=15mm, minimum height=7mm},
  gt/.style={draw, fill=black!6, minimum width=10mm, minimum height=6mm}]
  \node[Dn] (D0)   at (0,1.7)    {$D_{\ell-1}$};
  \node[op] (norm) at (2.6,1.7)  {RMSNorm};
  \node[op] (f)    at (5.4,1.7)  {$f_\ell$\,(attn/MLP)};
  \node[op] (upd)  at (8.6,1.7)  {erase\,+\,write};
  \node[Dn] (D1)   at (11.4,1.7) {$D_{\ell}$};
  \node[Cn] (C0)   at (0,0)      {$C_{\ell-1}$};
  \node[Cn] (C1)   at (11.4,0)   {$C_{\ell}$};
  \node[gt] (g)    at (11.4,0.85){gate $c_\ell$};
  \draw[->] (D0) -- (norm);
  \draw[->] (C0) to[out=25,in=215] node[pos=0.6,below=-1pt]{\scriptsize$\lambda_\ell$} (norm.south west);
  \draw[->] (norm) -- (f);
  \draw[->] (f) -- node[above]{\scriptsize$y_\ell$} (upd);
  \draw[->] (D0) to[out=65,in=115] (upd);          
  \draw[->] (upd) -- (D1);
  \draw[->] (D1) -- (g);
  \draw[->] (g) -- (C1);
  \draw[->] (C0) -- node[below]{\scriptsize append-only (protected)} (C1);
  \draw[->, very thick] (C1) -- ++(2.0,0) node[right]{Unembed};
  \node[font=\itshape\footnotesize, anchor=east] at (-0.9,1.7) {Deliberation};
  \node[font=\itshape\footnotesize, anchor=east] at (-0.9,0)   {Commitment};
\end{tikzpicture}
\caption{\textbf{Ledger Residuals.} The residual stream is split into a mutable Deliberation
stream $D$, an erasable scratchpad, and a protected, append-only Commitment stream $C$, the only
stream the unembedding decodes. Each sublayer reads from $D$ with a little of $C$ mixed in,
updates $D$ by erasing and then writing, and may promote the result into $C$ through a
one-directional commit gate $c_\ell$.}
\label{fig:method}
\end{figure*}

\subsection{Architecture}
Ledger Residuals replace the single residual stream with two streams that play different roles, shown
in Figure~\ref{fig:method}. The first is the Deliberation stream $D$, a scratchpad. Every sublayer
may write to it and, just as importantly, erase from it, so that stale intermediate content can
actually be removed instead of merely buried under later writes. The second is the Commitment stream
$C$, a protected ledger. Sublayers may append to it through a gate but can never overwrite it, and it
is the only stream the unembedding reads. Put plainly, $D$ is where the model works and $C$ is where
it records the answer it will be judged on. The whole point of the design is to give the model the
stable, content-independent place to keep its readout that, under the artifact hypothesis, it would
otherwise have to improvise for itself.

Concretely, we carry two width-$d$ tensors per token, initialized to $D_0=\mathrm{embed}(x)$ and
$C_0=\mathbf{0}$. For each sublayer $f_\ell$, which is either an attention block or an MLP block, the
two streams update as
\begin{align}
u_\ell &= \mathrm{RMSNorm}\!\left(D_{\ell-1} + \lambda_\ell \odot C_{\ell-1}\right), \quad y_\ell=f_\ell(u_\ell)\\
D_\ell &= D_{\ell-1} - \beta^e_\ell\odot(k_\ell^\top D_{\ell-1})\,k_\ell + \beta^w_\ell\odot g(y_\ell)\\
C_\ell &= C_{\ell-1} + c_\ell \cdot P(D_\ell).
\end{align}
The first line is the read. The sublayer sees the deliberation state with a small, learned amount of
the commitment state mixed in, normalized before use. The second line is the deliberation update,
written as a delta rule. Its first term, $\beta^e_\ell\odot(k_\ell^\top D_{\ell-1})k_\ell$, erases the
part of $D$ that lies along a learned direction $k_\ell$, and its second term writes the new content
$g(y_\ell)$, with the channel-wise gates $\beta^e_\ell$ and $\beta^w_\ell$ setting how much is erased
and how much is written. The third line is the commitment update. A scalar commit gate
$c_\ell\in(0,1)$, biased to stay closed in early layers and open in later ones, decides how much of the
current deliberation state to append to the ledger through a learned map $P$. We keep the maps $g$ and
$P$ low-rank and initialize them to the identity, which holds the extra parameters to about seven
percent at the 160M scale. The coupling between the streams runs in only one direction by
construction: the commitment stream influences deliberation only weakly, through the small read
coefficient $\lambda_\ell$, and is never written into by it. Decoding happens from the commitment
stream alone, with logits formed as $\mathrm{Unembed}(\mathrm{RMSNorm}(C_L + \gamma D_L))$ and the
mixing weight $\gamma$ annealed to zero over training, so that by the end the answer has to be read
out of $C$ by itself.

\subsection{Reduction to a standard transformer}
Because the comparison we care about is between a standard transformer and one with this
factorization, the factorization has to add nothing on its own, and it does not. Close the erase gate,
open the write gate fully, set $g$ to the identity, allow no commitment, and decode with $\gamma=1$,
and the deliberation update collapses to ordinary additive accumulation while the readout reduces to
$\mathrm{RMSNorm}(D_L)$, which is exactly a standard pre-norm transformer. We check this numerically:
at that setting the two models produce logits that differ by less than $10^{-3}$. Any difference we
see in the experiments therefore comes from the residual factorization itself and not from extra
capacity.

\subsection{What we measure}
To describe the outliers in each channel we report a few statistics, all computed per sublayer on a
validation batch, and we deliberately keep apart two things that are easy to conflate: how large an
outlier is, and how structured it is. For magnitude we use the excess kurtosis of the activations, a
standard heavy-tail measure that climbs when a few values dominate. For structure we use what we call
the fixed-dimension ratio. For each dimension we take the mean absolute activation across tokens, then
report the largest of these divided by the median. This is large exactly when a few input-independent
dimensions are persistently big, which is the defining signature of a massive activation. We also
track the single most prominent dimension in two ways: its persistence, the fraction of layers in
which it is the top outlier, and its start-token concentration, its magnitude on the first token
divided by its magnitude over all tokens, where a value above one means the outlier sits
preferentially on the start token. Finally, to see whether any of this structure has a practical
consequence, we apply post-training weight quantization to eight and four bits and measure the change
in validation loss.

\section{Experimental Setup}
We train decoder-only transformers on the FineWeb-Edu corpus at two scales, 160M and 290M parameters,
both using the GPT-2 byte-pair vocabulary. Within each scale we match every configuration on
parameters, data, optimizer, and token budget, so that the residual operator is the only thing that
changes. We train to convergence and report every statistic at matched validation loss. This matching
matters, because an outlier that showed up only because one model happened to be trained a little
better or worse would tell us nothing about the residual factorization, and matching the loss takes
that confound off the table.

We compare the four configurations summarized in Table~\ref{tab:configs}. Vanilla is a standard
pre-norm transformer. Suppress adds QK-normalization, which is known to reduce these outliers, and
stands in for the artifact-side baseline. Ledger is our architecture with a light commit-sparsity
penalty. The fourth, ledger-hs, is the same architecture with a six times stronger commit-sparsity
penalty, which we run at 290M to ask whether forcing the commitment channel to be sparse keeps an
outlier from forming in it. For vanilla and suppress the decoded channel is the ordinary residual
state, and for the two ledger configurations it is the commitment stream.

\begin{table}[t]\centering\small
\begin{tabular}{@{}lll@{}}
\toprule
Config & Residual operator & Decodes from \\
\midrule
vanilla    & standard pre-norm        & $h$ \\
suppress   & pre-norm $+$ QK-norm     & $h$ \\
ledger     & scratch/commitment       & $C$ \\
ledger-hs  & ledger, $6\times$ sparsity & $C$ \\
\bottomrule
\end{tabular}
\caption{The four configurations. All else is matched.}
\label{tab:configs}
\end{table}

\section{Results}
\label{sec:results}
\subsection{The outliers emerge, and QK-normalization suppresses them}
The phenomenon we set out to test is plainly present in our models, and it gets stronger with scale.
At 160M, where all configurations reach a validation perplexity of about 41, the vanilla model
develops a fixed-dimension outlier with an excess kurtosis of 81 and a fixed-dimension ratio of 34,
and its dominant dimension is the top outlier in 54 percent of layers. At 290M, where perplexity is
about 56, the same outlier becomes more entrenched still, with its dominant dimension topping the list
in 92 percent of layers. The suppress configuration behaves just as prior work would predict [1, 3].
QK-normalization lowers the fixed-dimension ratio at both scales, for instance from vanilla's 28.3 to
6.8 at 290M, and at 290M it also wipes out the start-token concentration, dropping it from 1.01 to
0.13. So far everything is as expected, which sets up the question the rest of the section answers.

\subsection{The protected channel does not stay clean}
\begin{figure}[t]\centering
\includegraphics[width=\linewidth]{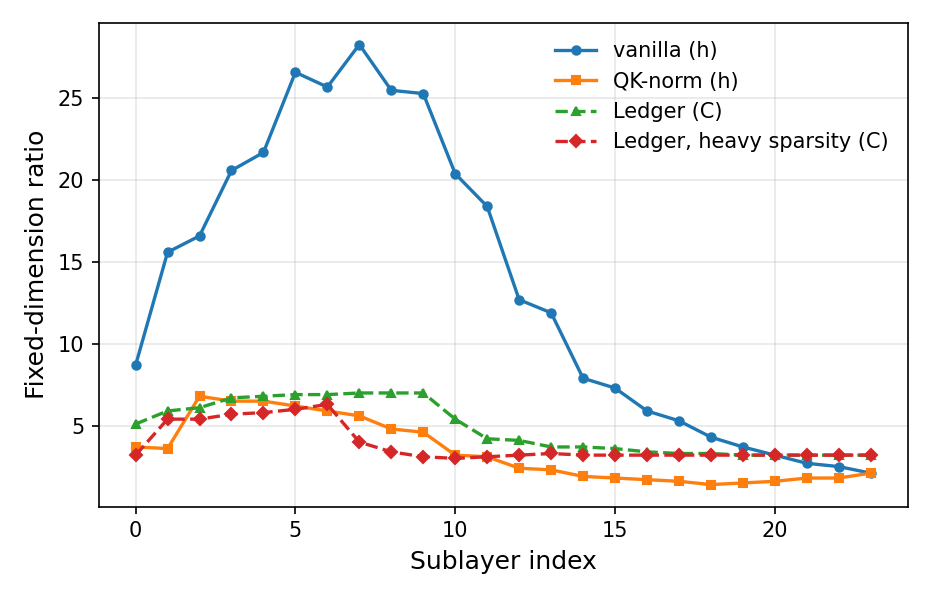}
\caption{Fixed-dimension outlier ratio by sublayer for the decoded channel at 290M (matched
loss). Ledger lowers the outlier magnitude relative to vanilla but does not remove the
fixed-dimension structure reported in Table~\ref{tab:main}.}
\label{fig:ratio}
\end{figure}

\begin{figure}[t]\centering
\includegraphics[width=\linewidth]{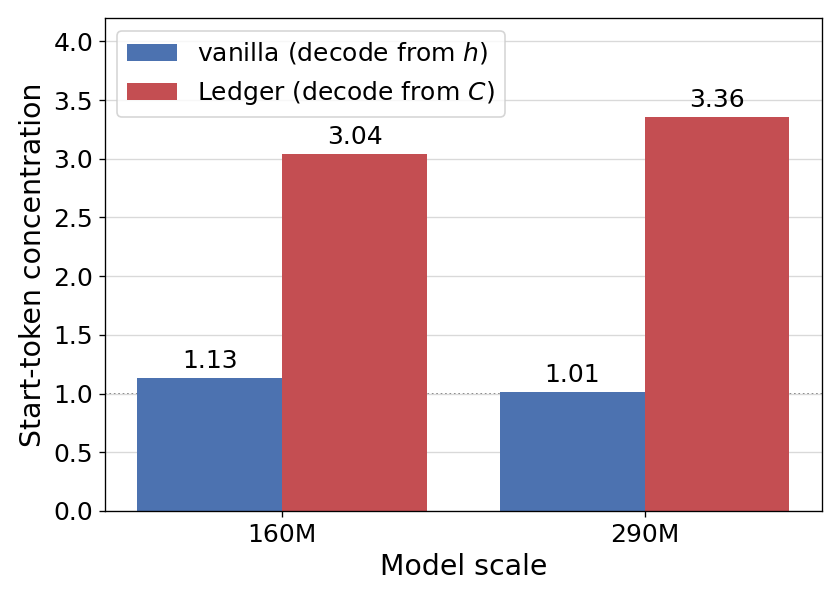}
\caption{The reconstruction holds across scale. At both 160M and 290M the decoded channel carries a
strongly start-token-concentrated outlier under Ledger (which decodes from $C$) and a near-neutral
value under a standard transformer (which decodes from $h$); the dotted line marks concentration 1.}
\label{fig:scale}
\end{figure}

Giving the model a protected, decode-only channel does not get rid of the outlier. Instead the
decoded commitment stream rebuilds it, as Table~\ref{tab:main} shows. At 290M
the commitment channel does carry a lower-magnitude outlier than vanilla, with a fixed-dimension ratio
of 7.0 against vanilla's 28.3 (Figure~\ref{fig:ratio}). But that outlier sits on the very same
dimension as the model's own scratch stream, and it is strongly concentrated on the start token, with
a concentration of 3.36 against vanilla's 1.01 and suppress's 0.13. The channel the model reads its
answer from has rebuilt the canonical start-token massive activation. Rather than being the cleanest
of the four channels, it is the only one that keeps a strongly start-token-locked readout feature. The
160M run tells the same story, with the commitment channel rebuilding the outlier at a fixed-dimension
ratio of 13 and a start-token concentration of 3.04, once again on the same dimension as the scratch
stream. As Figure~\ref{fig:scale} shows, the decoded channel's start-token concentration stays high
under Ledger at both scales and close to one under vanilla.

\begin{table}[t]\centering\small
\setlength{\tabcolsep}{4pt}
\begin{tabular}{@{}lcccc@{}}
\toprule
Channel (290M) & dim & persist. & ratio & BOS \\
\midrule
vanilla $h$            & 828 & 0.92 & 28.3 & 1.01 \\
suppress $h$           & 550 & 0.50 & 6.8  & 0.13 \\
\textbf{ledger $C$}    & 22  & 0.58 & 7.0  & \textbf{3.36} \\
\textbf{ledger-hs $C$} & 22  & \textbf{0.96} & 6.3  & \textbf{4.69} \\
\midrule
ledger $D$ (scratch)   & 22  & 0.50 & 17.8 & 2.63 \\
\bottomrule
\end{tabular}
\caption{Decoded-channel outlier signature at $290$M (matched loss); metrics are defined in
Section~\ref{sec:method}. The protected channel $C$ rebuilds the same dominant dimension as
scratch $D$, with strong start-token concentration (BOS $>1$).}
\label{tab:main}
\end{table}

\subsection{Forcing sparsity makes it stronger, not weaker}
A reasonable objection is that the commitment channel rebuilds the outlier only because it is free to
commit as often as it likes, and that forcing it to be sparse would stop a persistent fixed-dimension
feature from forming in the first place. We test this with ledger-hs, which applies a six times
stronger commit-sparsity penalty, and the result is the opposite of that prediction. The rebuilt
outlier gets stronger rather than weaker. Its dominant-dimension persistence climbs from 0.58 to 0.96
and its start-token concentration from 3.36 to 4.69 (Table~\ref{tab:main}). Allowed fewer commitments,
the model does not drop the start-token feature; it pours its scarce commitments into it, concentrating
the feature into a single, nearly ever-present dimension. The architecture cannot squeeze the feature out, and the added pressure only intensifies it.

\subsection{The relocation has no measurable consequence}
None of this shows up as a change in capability, at least not by the measures we have. All four
configurations land at the same validation perplexity at each scale, within a narrow band of 55.4 to
57.7 at 290M. Weight quantization tells the same story. Going to eight bits is lossless, going to four
bits costs about 1.3 percent of perplexity, and those costs are indistinguishable across the four
configurations. The one thing Ledger does accomplish, lowering the magnitude of the decoded outlier, has no effect on these measures, and QK-normalization does the same more simply anyway.

\section{Discussion}
\subsection{Why does the protected channel rebuild the feature?}
The artifact hypothesis predicts the feature should vanish once the model has a clean place to keep its
answer. What we see instead is the feature rebuilt. The simplest explanation is that the start-token,
fixed-dimension feature is something the model actively relies on, a content-independent channel that
doubles as an attention sink and a bias term, and that the model will reconstruct it in whatever
representation it ends up decoding from. A dedicated, protected channel does not remove the need for
the feature; it only moves where the feature gets built. The sparsity result is this conclusion in its sharpest form: restricting how often the model may commit only deepens its reliance on the feature.

\subsection{Implications for the artifact-versus-function debate}
Earlier removal results, using explicit biases, QK-normalization, or rectified softmax variants,
establish that the magnitude of these outliers can be suppressed at little or no cost to language
modeling [1, 3, 9]. We agree on magnitude. What we add is a distinction the magnitude-only view
misses, namely that magnitude and structure are not the same thing. Even when the architecture hands
the model a clean alternative answer channel, the start-token, fixed-dimension structure comes back in
it. That favors a functional account, in the precise sense that the structure survives a direct
attempt to design it away. It does not tell us which function is responsible. The feature could be
serving mainly as an attention sink, as a bias term, as a gradient regulator, or as some mix of these,
and our experiments do not adjudicate between them.

\subsection{Limitations}
Our experiments are small by current standards. The largest model is 290M parameters, well below the
multi-billion-parameter regime where massive activations are most extreme, and in our models the
absolute magnitudes are modest and fade toward the final layer even though the start-token structure
persists across most of the network. We train one seed per configuration and use English web text
only. Our capability measures are validation perplexity and weight quantization, and we did not find a
task where the relocation of outlier magnitude makes a difference; a more demanding test would go
looking for one, perhaps through activation quantization or a downstream task that leans hard on the
answer representation. Finally, Ledger is one particular way of building a protected channel, and
another way might behave differently, though the result holding across two scales and two sparsity
settings makes that less likely.

\section{Conclusion}
We asked whether massive activations are a removable artifact of the residual stream's overloaded
role, and we tested it by giving the model an explicit, protected, decode-only channel for its answer.
The activations are not removed. Across 160M- and 290M-parameter models at matched loss, the model
rebuilds the canonical fixed-dimension, start-token outlier inside the protected channel, and
pressuring that channel to be sparse only intensifies the outlier. Massive activations are
architecturally robust in this sense: they re-emerge in whichever representation the model decodes
from. We read that as evidence on the functional side of the artifact-versus-function debate. To help
others build on this, we release Ledger Residuals and the activation-probe code so that the result can
be reproduced and pushed to larger scale.


\end{document}